# Opening the Black Box: An Explainable, Few-shot AI4E Framework Informed by Physics and Expert Knowledge for Materials Engineering

Haoxiang Zhang[1, *], Ruihao Yuan[2], Lihui Zhang[1], Yushi Luo[1], Qiang Zhang[1], Pan Ding[1], Xiaodong Ren[1], Weijie Xing[1], Niu Gao[1], Jishan Chen[1], Chubo Zhang[1]

[1] Science and Technology on Advanced High Temperature Structural Materials Laboratory, Beijing Institute of Aeronautical Materials, Beijing, 100095, China

[2] State Key Laboratory of Solidification Processing, Northwestern Polytechnical University, Xi'an, 710072, China

Authors to whom correspondence should be addressed: hxzhang0305@163.com (H.Z.)

The industrial adoption of Artificial Intelligence for Engineering (AI4E) faces two fundamental bottlenecks: scarce high-quality data and the lack of interpretability in black-box models—particularly critical in safety-sensitive sectors like aerospace. We present an explainable, few-shot AI4E framework that is systematically informed by physics and expert knowledge throughout its architecture. Starting from only 32 experimental samples in an aerial K439B superalloy castings repair welding case, we first augment physically plausible synthetic data through a three-stage protocol: differentiated noise injection calibrated to process variabilities, enforcement of hard physical constraints, and preservation of inter-parameter relationships. We then employ a nested optimization strategy for constitutive model discovery, where symbolic regression explores equation structures while differential evolution optimizes parameters, followed by intensive parameter refinement using hybrid global-local optimization. The resulting interpretable constitutive equation achieves 88% accuracy in predicting hot-cracking tendency. This equation not only provides quantitative predictions but also delivers explicit physical insight, revealing how thermal, geometric, and metallurgical mechanisms couple to drive cracking-thereby advancing engineers' cognitive understanding of the process. Furthermore, the constitutive equation serves as a multi-functional tool for process optimization and high-fidelity virtual data generation, enabling accuracy improvements in other data-driven models. Our approach provides a general blueprint for developing trustworthy AI systems that embed engineering domain knowledge directly into their architecture, enabling reliable adoption in high-stakes industrial applications where data is limited but physical understanding is available.

**Key words**：AI4E, Interpretable AI, Machine Learning, Symbolic Regression, Superalloy, Aircraft engine castings

# INTRODUCTION

The translation of advanced materials from laboratory discovery to industrial application remains a pivotal challenge in materials engineering. While initiatives such as the Materials Genome Initiative have significantly accelerated the design of novel materials through composition-structure-property relationships[1-6], downstream engineering processes, such as manufacturing parameter optimization and quality prediction, still rely heavily on empirical methods or numerical simulations[7-11]. This reliance stems from the inherent complexity of industrial processes and the generally underdeveloped AI infrastructure—both in expertise and equipment—within the materials manufacturing sector, leading to inadequate AI empowerment in practical materials engineering applications[12].

The convergence of data-driven AI with modern manufacturing technology has established Artificial Intelligence for Engineering (AI4E) as a transformative paradigm in materials engineering[13-15]. This approach is increasingly being applied to core industrial tasks, including performance prediction, parameter optimization, and computational simulation[16-18]. However, as AI4E sees growing preliminary adoption, its broader and deeper implementation is constrained by two fundamental issues that limit its full potential. The first is the interpretability problem. Conventional data-driven AI models often function as "black boxes" offering limited physical insight and eroding engineer trust, which hinders both scientific discovery and high-stakes industrial adoption where reliability is paramount[19]. This is especially critical in sensitive and safety-first sectors such as aerospace, defense, and nuclear energy, where model decisions must be transparent, verifiable, and physically grounded to ensure operational reliability and security. Although recent explainable AI techniques, such as Kolmogorov-Arnold Networks (KANs) and genetic programming (GP) symbolic regression methods can construct interpretable formulas for scientific problems with highly correlated parameters[20-22], they prove inadequate for complex materials engineering problems marked by high-dimensional, multi-scale parameters and incomplete domain knowledge. The second barrier is data scarcity. Data-driven AI models generally require large volumes of labeled training data, yet industrial materials production lines usually generate only sparse, fragmented datasets due to limited sensor coverage and low digitalization levels. This issue is particularly acute for high-value products, where the available datasets are even scarcer due to cost constraints, compounding the challenge. While transfer learning provides a theoretical means to incorporate knowledge from related tasks[23, 24], it offers limited utility in bespoke or niche industrial settings where no relevant source domains exist. Moreover, although data augmentation is mature in fields such as computer vision[25-27], analogous methods for enriching numerical process parameters in materials engineering remain underdeveloped, creating a critical bottleneck for model training and validation.

Under these constraints, it is essential to recognize that materials engineers often possess deep domain expertise and a firm grasp of the physical principles that govern production processes. This knowledge represents a valuable—and often underutilized—resource. In the absence of sufficient data and inherently interpretable algorithms, embedding such expert knowledge and physical rules into AI frameworks offers a viable path to improve both transparency and data efficiency.

To address these challenges, this work introduces a physics- and knowledge-informed AI4E framework that supports explainable modeling under limited data conditions. Our approach systematically integrates symbolic regression with data augmentation, both of which are structured by physics and expert insight. Starting from limited experimental data, we first generate physically plausible datasets through a three-stage protocol incorporating differentiated noise injection, enforcement of hard physical constraints, and preservation of inter-parameter relationships. We then employ a nested optimization strategy for constitutive model discovery, in which symbolic regression explores equation structures while differential evolution optimizes parameters, followed by intensive refinement via hybrid global-local optimization. The resulting interpretable constitutive equation not only provides quantitative predictions but also delivers explicit physical insight—revealing how thermal, geometric, and metallurgical mechanisms couple to drive cracking—thereby advancing engineers' cognitive understanding of the process. Furthermore, the equation serves as a multi-functional tool for process optimization and high-fidelity virtual data generation, offering a practical route toward trustworthy AI in industrial materials processing. This methodology is also relevant to other engineering domains facing similar challenges of data scarcity and the need for model interpretability.

## RESULTS

Our proposed AI4E framework successfully transforms a severely limited set of experimental observations into a robust, interpretable, and high-performance predictive model through a meticulously designed, physics-informed pipeline (Fig. 1). This process, demonstrated on the critical problem, progresses sequentially through three core stages: physics-informed data augmentation, automated discovery of an explainable constitutive equation, and hybrid model enhancement. Beginning with few-shot experimental samples, we first expanded the dataset through physics-constrained augmentation while simultaneously decomposing the complex engineering problem into fundamental physical terms based on domain knowledge and expert insight. We then employed a two-stage optimization strategy to explore the mathematical relationships among these physical terms, constructing an interpretable constitutive equation. Finally, the discovered equation was deployed for defect prediction, process optimization, and high-fidelity virtual data generation, enabling subsequent enhancement of hybrid models. Each step is rigorously governed by physical principles and domain expertise, ensuring that the final model is both data-efficient and physically meaningful.

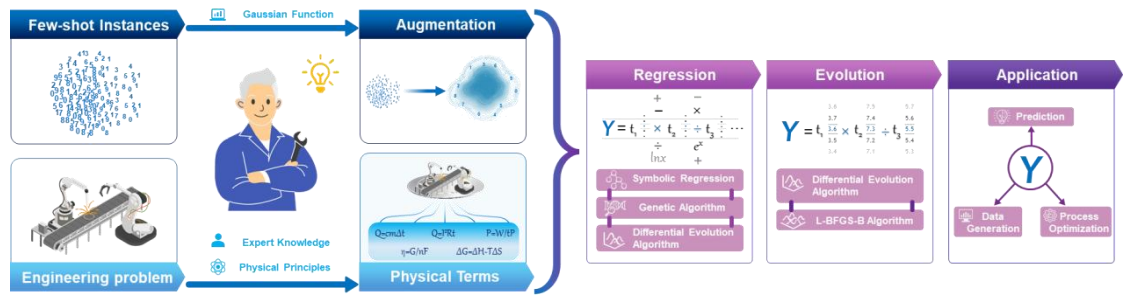

**Figure 1** Schematic overview of the AI4E framework. The workflow progresses through three sequential stages: (1) physics-constrained augmentation of limited experimental data coupled with decomposition of the engineering problem into fundamental physical terms; (2) two-stage discovery of mathematical relationships among physical terms to construct an interpretable constitutive equation; and (3) application of the constitutive equation for defect prediction, process optimization, and virtual data generation to enhance hybrid models.

We selected a representative and industrially critical scenario from materials engineering to validate our framework: hot-cracking prediction during the repair welding of aerial K439B nickel-based superalloy castings[28-30]. This scenario was chosen because it quintessentially embodies the core challenges that our AI4E framework is designed to address: a high-dimensional parameter space governed by intricate thermo-mechanical-metallurgical interactions, the critical demand for model reliability and interpretability in safety-sensitive applications like aerospace, and the extreme data scarcity inherent to costly and time-intensive industrial experiments, making it an ideal testbed for our few-shot, explainable AI4E methodology.

**Physics-Informed Data Augmentation Constructs a Physically Plausible Parameter Space**

This repair welding process is directly affected by the following 9 factors, as shown in Table 1. Faced with an initial dataset of only 32 samples—a number that reflects the practical constraints of costly and time-intensive industrial experimentation — we first tackled the fundamental challenge of data scarcity. Our approach went beyond conventional statistical interpolation, employing instead a knowledge-guided data augmentation protocol engineered to systematically expand the dataset while rigorously conforming to physical laws.

Table 1  nine key physical parameters affecting the results of soldering repair defects

| Number | Parameter | Meaning | Unit | Noise Amplitude ($\sigma$) |
|---|---|---|---|---|
| 1 | $I$ | The current used during welding | A | ±1.5 A |
| 2 | $\tau$ | The duration used for each welding pass | s | ±1.5 s |
| 3 | $t_b$ | Base metal thickness | mm | ±0.3 mm |
| 4 | $d_w$ | The deepest depth of the welding point | mm | ±0.5 mm |
| 5 | $A_w$ | The area of the welding point | mm | ±10 mm² |
| 6 | $T_i$ | Interpass temperature of welding point | °C | ±20°C |
| 7 | $T_p$ | Preheating temperature before welding | °C | ±15°C |
| 8 | $t_{500}$ | The duration from the welding completion until the welding point cooling down to 500°C | s | ±5 s |
| 9 | $d$ | Welding wire diameter | mm | ±0.1 mm |

The procedure began with differentiated Gaussian noise perturbation[31]. Unlike uniform noise addition, each of the nine welding parameters was perturbed with a unique noise amplitude, as specified in Table 1. These amplitudes were meticulously calibrated according to operational tolerances and empirical engineering knowledge, thereby realistically capturing inherent process fluctuations and generating a cloud of candidate data points.

This stochastically enriched dataset was subsequently processed through a hard physical constraint layer, founded on deep domain expertise. This layer enforced deterministic rules to ensure all generated samples remained physically admissible. Perturbations to interdependent parameters were co-adjusted to maintain process consistency—for instance, an increase in wire diameter triggered a commensurate increase in welding current to preserve arc stability. Each parameter was then clipped to its physically admissible range, as defined by material properties and process limitations.

By integrating this multi-layered constraint system with carefully calibrated noise amplitudes, we ensured that all synthesized samples retained the original outcome labels (cracked/not cracked). The original 32 samples were thus transformed into a robust training ecosystem of 32,000 physically plausible data instances, effectively encapsulating process variance within well-defined physical bounds and establishing a reliable foundation for subsequent model discovery. Fig. 2 shows the distribution of the initial data as well as the augmented data. It is seen that the data diversity is clearly expanded. Such augmentation can benefit the following surrogate model as new data points convey additional information.

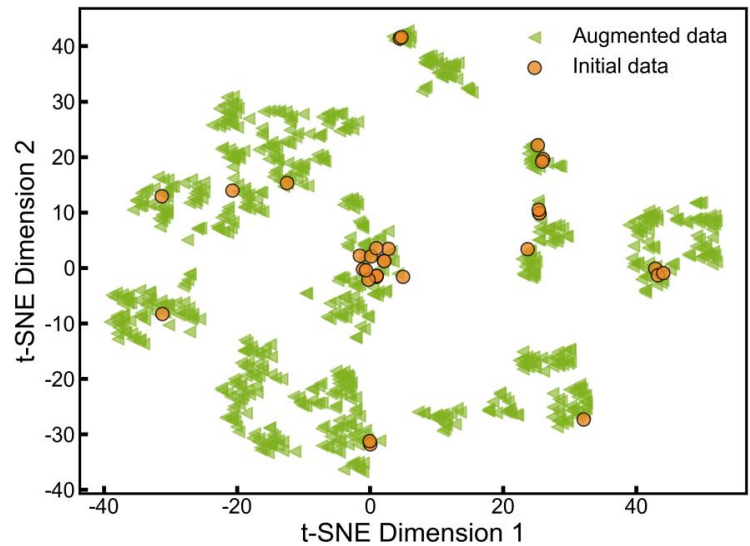

Figure 2. The distribution of the initial data and the augmented data base on tSNE dimension reduction method, the data diversity is largely expanded [32-34].

**Symbolic Regression Discovers an Explainable Constitutive Model Embedding Domain Physics**

With the augmented dataset providing a rich exploration of the parameter space, we proceeded to the core task of distilling an explainable model. The derivation of a physically interpretable yet accurate constitutive model from a high-dimensional parameter space presents a significant computational challenge. To address this, we implemented a structured methodology comprising two complementary phases: an initial discovery phase that simultaneously identifies both the equation structure and a preliminary parameter set, followed by an intensive refinement phase that optimizes these parameters to achieve maximum predictive accuracy.

The foundation of our approach lies in engineering a set of fundamental physical terms that encapsulate the key mechanisms governing hot-cracking susceptibility. These six composite terms, formulated from domain expertise and physical first principles, are detailed in Table 2, and associated physical constants are presented in Table 3. This step transforms the raw, high-dimensional parameter space into a lower-dimensional space of physically meaningful quantities, thereby constraining the subsequent symbolic search to physically plausible relationships. It is important to note that the primary objective of this constitutive model is to predict a probability of cracking, not to compute a specific physical quantity with strict dimensional homogeneity. Therefore, to eliminate the influence of differing units and magnitudes among the various physical terms, each term was raised to a dimensionless exponent ($\alpha$, $\beta$, $\gamma$, etc.) during the symbolic regression process. This approach allows the model to flexibly capture the relative scaling and interplay between different physical effects without being constrained by dimensional consistency, focusing purely on their functional relationship to the cracking probability index (CPI).

**Table 2: Engineered Physical Terms for Constitutive Modeling**

| Physical Term | Mathematical Formulation | Physical Interpretation | Source |
|---|---|---|---|
| T1: Current Heat Input | $(\frac{I \cdot U \cdot t}{\sqrt{A}})^{\alpha}$ $= (\frac{4 \cdot I \cdot U \cdot \sqrt{A} \cdot h}{\Pi \cdot d^2 \cdot wire\_feed\_speed(d)})^{\alpha}$ | Represents the energy input from the arc, modified by wire feed dynamics. Index $\alpha$ is applied to eliminate the quantitative factors. | Derived from a existing physical formulas. $Heat\ Input = \frac{I \cdot U}{v} = \frac{I \cdot U \cdot t}{L} \approx \frac{I \cdot U \cdot t}{\sqrt{A}}$[35] $t \approx \frac{4 \cdot A \cdot h}{\Pi \cdot d^2 \cdot wire\_feed\_speed(d)}$ |
| T2: Geometric | $(\frac{d_w}{t_b})^{\beta}$ | Characterizes stress concentration via the weld | Derived from the expertise of experts. |

| Physical Term | Mathematical Formulation | Physical Interpretation | Source |
|---|---|---|---|
| Constraint | | depth to base thickness ratio. Index β is applied to eliminate the quantitative factors. | |
| T3: Thermal Stress | $(\theta_1(\delta - T_i) + \theta_2(\delta - T_p))^\gamma$ | Captures the combined driving force for thermal stresses from interpass and preheat temperatures relative to the maximum temperature δ of the welding points. Index γ is applied to eliminate the quantitative factors. | Derived from a existing physical formulas. Thermal Stress $=\alpha\, E\, (T - T_0)$ [36] |
| T4: Heat Accumulation | $(T + \triangle T)^\varepsilon = (T_1 + \dfrac{I \cdot U \cdot \tau}{t_b \cdot \rho \cdot c \cdot \Pi \cdot r^2})^\varepsilon$ | The temperature rise from both interpass temperature and the diffusion of newly input arc heat. Index ε is applied to eliminate the quantitative factors. | Derived from a existing physical formulas and expertise of experts. $\triangle T = \dfrac{Q}{m \cdot c}$ [37] |
| T5: Cooling Rate | $\dfrac{(\delta - 500)^\zeta}{t_{500}}$ | Quantifies the critical cooling rate through the solidification temperature | Derived from expertise of experts. |

| Physical Term | Mathematical Formulation | Physical Interpretation | Source |
|---|---|---|---|
| | | range. Index ζ is applied to eliminate the quantitative factors. | |
| T6: Wire Diameter | $(\frac{1}{d})^{\eta}$ | Accounts for the influence of wire diameter on melt pool geometry and solidification. Index η is applied to eliminate the quantitative factors. | Derived from expertise of experts. |

**Table 3  Associated Constants and Functions**

| Name | Symbol | Value / Definition |
|---|---|---|
| Welding Voltage | $U$ | 12 V |
| Base Material Density | $\rho$ | 8.16 g/cm³ |
| Base Material Specific Heat Capacity | $c$ | 0.75 J/(g · K) |
| Arc Heat Affect Zone Radius | $r$ | 50.0 mm |
| Wire Feed Speed | $wire\_feed\_speed(d)$ | Function of (d) if d > 2.5: return 2.0 mm/s elif 1.8 <= d <= 2.5: return 3.0 mm/s else: return 4.0 mm/s |

Subsequently, a nested optimization strategy was employed to uncover the operational relationship between these physical terms. A genetic programming-based symbolic regression algorithm explored the combinatorial space of possible equation structures, such as +, -, ×, ÷, lnx, e$^x$, while a differential evolution optimizer, acting as an inner loop, tuned the associated parameters—scaling factor K, exponents α-η, the maximum temperature δ, and coefficients $θ_1$ and $θ_2$—within predefined physically-plausible bounds[38, 39]. To evaluate the predictive performance of each candidate constitutive equation, the computed CPI was converted into a cracking probability via a logistic function: $P_{crack} = \frac{1}{(1+e^{(-k(CPI-CPI_{crit})})}$ with k=2 and $CPI_{crit}$=0.6. A sample was predicted as "cracked" if its $P_{crack} > 0.5$, and "not cracked" otherwise. Model fitness was evaluated using a comprehensive accuracy metric, defined as the weighted average of accuracy on the validation set (50%) and accuracy on the original 32 experimental samples (50%). This balanced metric ensures that discovered models maintain strong performance on both the augmented data and the original ground-truth experiments. This co-discovery process identified the constitutive model CPI = K * (($T_1$ * $T_4$) + ($T_2$ * $T_6$) + ($T_3$ * $T_5$)) as the most promising constitutive model with preliminary parameters were: K = 6.292400e$^{-2}$, α = 0.000152, β = 3.463110, γ = 5.521969, δ = 1380.534054, ε = 0.332720, ζ = 0.792024, η = 0.215927, $θ_1$ = 1.562374e$^{-9}$, $θ_2$ = 1.014778e$^{-9}$, achieving 84% comprehensive accuracy.

With the functional form of the constitutive equation established, we focused on a rigorous refinement of its parameters to enhance predictive performance. This phase treated the discovered equation structure as a fixed, interpretable scaffold. We initiated a global search using a differential evolution algorithm, which operated over deliberately expanded parameter bounds defined as 0.1-10.0 times the initial values to avoid local optima. The most promising candidate from this global search was subsequently used to initialize a local gradient-based refinement using the L-BFGS-B algorithm[40], ensuring convergence to a high-precision optimum.

This hybrid optimization protocol yielded a refined parameter set that markedly improved the model's predictive capability: K = 5.229864e$^{-2}$, α = 0.000426, β = 2.233428, γ = 4.329339, δ = 1437.709817, ε = 0.362319, ζ = 0.383383, η = 1.898158, $θ_1$ = 3.055194e$^{-9}$, $θ_2$ = 2.306818e$^{-9}$, as shown in Fig 3. The finalized constitutive model, embodying this refined parameter set, achieved a comprehensive accuracy of 88%.

**Final constitutive equation for the welding crack probability index (CPI)**

$$CPI = K \times \left( \begin{array}{l} \left[\left(\frac{4 \cdot I \cdot U \cdot A_w \cdot d_w}{\sqrt{A_w} \cdot \pi \cdot d^2 \cdot v_f}\right)^\alpha \times \left(T_i + \frac{I \cdot U \cdot \tau}{t_b \cdot \rho \cdot c \cdot \pi \cdot r^2}\right)^\varepsilon\right] \\ + \left[\left(\frac{d_w}{t_b}\right)^\beta \times \left(\frac{1}{d}\right)^\eta\right] \\ + \left[(\theta_1 \cdot (\delta - T_i) + \theta_2 \cdot (\delta - T_p))^\gamma \times \left(\frac{\delta - 500}{t_{500}}\right)^\zeta\right] \end{array} \right) = 0.05229864 \times \left( \begin{array}{l} \left[\left(\frac{4 \cdot I \cdot U \cdot A_w \cdot d_w}{\sqrt{A_w} \cdot \pi \cdot d^2 \cdot v_f}\right)^{0.000426} \times \left(T_i + \frac{I \cdot U \cdot \tau}{t_b \cdot \rho \cdot c \cdot \pi \cdot r^2}\right)^{0.362319}\right] \\ + \left[\left(\frac{d_w}{t_b}\right)^{2.233428} \times \left(\frac{1}{d}\right)^{1.898158}\right] \\ + \left[(3.055194 \times 10^{-9} \cdot (1437.709817 - T_i) + 2.306818 \times 10^{-9} \cdot (1437.709817 - T_p))^{4.329339} \times \left(\frac{937.709817}{t_{500}}\right)^{0.383383}\right] \end{array} \right)$$

**Figure 3** The Final constitutive equation for the welding crack probability index, and the highlighted letters represent the key physical parameters, and the cracking probability was calculated via a logistic function: $P_{crack} = \frac{1}{(1+e^{(-k(CPI-CPI_{crit})})}$ with k=2 and $CPI_{crit}$=0.6.

This stepwise strategy—decoupling the discovery of an interpretable equation structure from its subsequent parameter refinement—proved highly effective. It yielded a transparent model that not only delivers robust predictive performance but, more importantly, encapsulates and reveals the core physical mechanisms at play. The derived constitutive equation, CPI = K * (($T_1$ * $T_4$) + ($T_2$ * $T_6$) + ($T_3$ * $T_5$)), possesses a lucid physical interpretation. It synthesizes three primary driving mechanisms for hot-cracking: the combined effect of heat input and accumulation ($T_1$ * $T_4$), the interaction between geometric constraint and wire diameter influence ($T_2$ * $T_6$), and the coupling of thermal stress and cooling rate ($T_3$ * $T_5$).

This structure is fundamentally explainable. Engineers are no longer presented with an inscrutable black-box prediction; instead, they can directly quantify how changes in key process parameters influence each distinct physical mechanism and trace their collective contribution to the overall cracking risk. For high-stakes applications like aero-engine component repair, this explicit, causal understanding is paramount. The model achieves high accuracy (88%) while being inherently trustworthy, as every prediction is auditable against physical first principles. Furthermore, it actively promotes engineering cognition by clearly demonstrating that hot-cracking is not governed by a single factor, but by the synergistic coupling of thermal, geometric, and metallurgical mechanisms, thereby guiding more insightful process design and troubleshooting.

**The Constitutive Equation Serves as a Multi-Functional Tool for Prediction, Optimization, and Data Generation**

The explainable constitutive model, derived directly from data and physics, stands as a self-sufficient predictive instrument. Its application is straightforward: engineers can input a vector of proposed welding parameters into the explicit equation and immediately receive a quantitative prediction of cracking propensity. This transparency is its defining advantage. The model's 88% accuracy demonstrates robust predictive capability, successfully capturing the dominant physical mechanisms governing hot-cracking in K439B.

Furthermore, by employing traversal or optimization algorithms across the parameter space, the model can efficiently identify process windows (combinations of parameters) where the predicted cracking probability is minimized (e.g., <10%), directly enabling robust process design.

The discovered constitutive equation is not merely a predictive model; it is a versatile engineering asset. A pivotal application of the constitutive equation is its use as a high-fidelity, physics-based data generator. By sampling the input space and computing the corresponding CPI and crack probability, the equation can produce a vast number of physically consistent "virtual samples." This capability is transformative in a data-scarce context.

We constructed a multi-source, weighted data ecosystem by amalgamating three distinct data streams, as shown in Fig 4:

**Original Experimental Data (N=32):** Assigned the highest weight (w=5.0) to anchor any subsequent model to empirical ground truth.

**Physics-Augmented Data (N=32,000):** Generated from the initial raw parameters via our noise-injection and constraint method, assigned a unit weight (w=1.0).

**Constitutive-Model-Generated Virtual Data (N=32,000):** The constitutive equation was deployed as a high-fidelity generator of virtual samples. It was used to perform targeted interpolation within the augmented dataset, specifically enriching parameter regions that remained data-sparse even after augmentation. This generated an additional 32,000 virtual samples, all stamped with the physical plausibility inherent to their source equation, also assigned a unit weight (w=1.0).

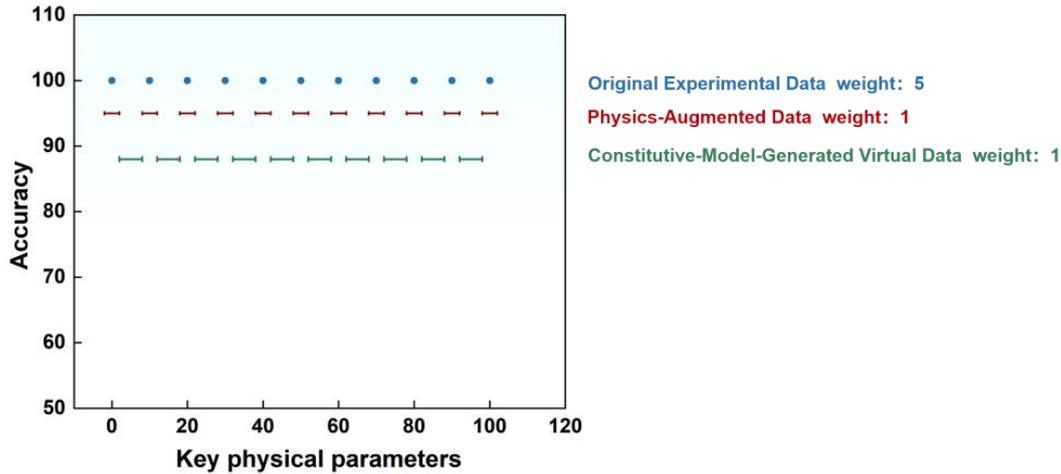

**Figure 4** Schematic diagram of multi-source, weighted data ecosystem. The three types of data constitute a complete data ecosystem.

This weighted dataset was used to train a deep neural network (DNN). The DNN's architecture (e.g., multiple hidden layers with non-linear activations) grants it the capacity to learn the complex, latent interactions between parameters that the constitutive equation's simpler form might miss. The result was a hybrid model that achieved a superior test accuracy of 97%. This significant improvement over the model trained only on the original data (76%) underscores the framework's power. The DNN effectively distills knowledge from the entire, physically-consistent data ecosystem, learning a more powerful predictive function while being anchored and regularized by the high-weight, real experimental data. This represents the culmination of the AI4E framework: the explainable model first overcomes data scarcity and injects physical knowledge, which in turn enables a high-performance data-driven model to reach its full potential. This sequential synergy ensures that the final high-accuracy predictor is built upon a foundation of physical understanding and trust.

## DISCUSSION

Our study establishes a physics- and knowledge-informed AI4E framework that systematically bridges the gap between data-driven artificial intelligence and fundamental engineering principles. This work makes several significant contributions to the field of materials engineering and beyond. First, we demonstrate that the integration of domain knowledge with symbolic regression enables the discovery of interpretable constitutive equations directly from limited experimental data—a capability that has remained elusive with conventional black-box approaches. Second, we develop a physics-constrained data augmentation methodology that generates physically plausible synthetic datasets, effectively addressing the critical bottleneck of data scarcity in industrial applications. Third, the resulting explainable model provides explicit physical insight by revealing how thermal, geometric, and metallurgical mechanisms interact to drive crack formation一thereby advancing engineers' cognitive understanding of the process beyond mere prediction. Fourth, we show that the resulting explainable models serve as versatile engineering assets that not only provide predictive capabilities but also enable process optimization and facilitate the generation of high-quality virtual data for enhancing more complex models.

The framework's architecture represents a paradigm shift from simply applying AI to engineering data toward constructing AI for engineering, where physical understanding guides both data generation and model discovery. This approach offers a pragmatic resolution to the long-standing trade-off between model interpretability and predictive performance. While purely physics-based models often struggle to capture complex phenomena without extensive calibration, and purely data-driven models lack transparency and physical consistency, our hybrid methodology leverages the strengths of both paradigms. The constitutive equation provides a physically-grounded foundation, while the neural network captures subtle, higher-order interactions that may be challenging to encapsulate in a compact analytical form.

A particularly noteworthy aspect of our methodology is its treatment of physical term engineering. Rather than presuming the definitive correctness of specific term formulations, we demonstrate how domain expertise一whether derived from first principles or empirical knowledge一can be systematically incorporated as symbolic regression priors to guide the model discovery process. This acknowledges the reality that engineering knowledge is often imperfect yet valuable, and provides a structured pathway for its integration with data-driven learning. The resulting models thus embody a synthesis of human understanding and machine discovery, making them both technically sound and practically useful for engineering decision-making.

The implications of this work extend beyond the specific case of welding crack prediction. The general framework of physics-informed data augmentation followed by symbolic regression and hybrid modeling can be adapted to numerous materials processing applications where data is scarce and physical understanding is partial but evolving. This includes applications in heat treatment optimization, solidification processing, and additive manufacturing parameter development, among others.

Several limitations of the current study warrant discussion. The computational demands of the symbolic regression process, while manageable for the present case, may become prohibitive for problems with higher dimensionality or more complex physical term structures. Future work could explore more efficient structure search algorithms or incorporate incremental learning approaches. Additionally, while our physical constraints effectively filtered implausible data regions, they may have inadvertently excluded some valid but uncommon process conditions. A more nuanced approach to constraint application, perhaps incorporating probabilistic rather than hard boundaries, might capture a broader range of feasible operating conditions.

Despite these limitations, the proposed framework offers a robust foundation for trustworthy AI in engineering applications. By maintaining physical consistency throughout the modeling pipeline — from data generation to equation discovery to final prediction — we establish a methodology that engineers can understand, validate, and trust. This transparency is particularly crucial in safety-critical applications where model reliability is paramount.

Looking forward, several research directions emerge from this work. The integration of more sophisticated physical priors, such as differential constraints or symmetry invariances, could further enhance the efficiency and accuracy of the model discovery process. Extending the framework to dynamic and multi-scale problems represents another important frontier. Finally, developing more intuitive interfaces for domain experts to interact with and guide the AI system could accelerate the adoption of these methodologies in industrial practice.

## MATERIALS AND METHODS

### Experimental Design and Data Acquisition

The primary objective of this research was to develop and validate a general-purpose, physics-informed AI4E framework for high-stakes engineering applications under data scarcity, selecting a critically challenging industrial problem: predicting hot-cracking during repair welding of K439B nickel-based superalloy aerial castings as the testbed.

To generate the foundational few-shot dataset, we conducted a systematic experimental campaign. Welding trials were performed on cast K439B specimens (100 mm × 50 mm × 1.5–20 mm) using K439B filler wire and a TIG welding system (Migatronic PI350). The process parameters were meticulously controlled and monitored with high-precision instrumentation: a Yokogawa WT1800 power analyzer for current and voltage, S-type platinum-rhodium thermocouples (±0.1°C) for temperature profiling, and manual stopwatch timing for cooling cycles. Post-process metallographic examination quantified key geometric features like weld depth and area. This rigorous approach yielded a high-quality but small dataset of N=32 experimental samples, each representing a unique combination of parameters within the industrial operational window and annotated with a binary crack/no-crack label.

**Physics and Expert Knowledge-Informed Data Augmentation**

To overcome the severe data scarcity, we designed a deterministic data augmentation protocol that moves beyond naive statistical interpolation by embedding domain knowledge directly into the data generation process. The algorithm operates in three sequential, rule-governed stages:

Differentiated Gaussian Noise Injection: The original dataset was perturbed by adding Gaussian noise with a mean ($\mu$) of zero. Critically, the noise amplitude, defined by its standard deviation ($\sigma$), was uniquely calibrated for each of the 9 process parameters based on their empirically documented measurement uncertainties and operational tolerances (e.g., $\sigma = 1.5$ A for welding current, $\sigma = 0.5$ mm for weld depth), as systematically outlined in Table 1.

Following the initial perturbation, the data was passed through a multi-layered system of hard constraints, implemented as deterministic rules to guarantee physical plausibility. This system acted as a filter and corrector, enforcing the following principles:

**Geometric Compatibility:** The depth of the weld bead was constrained to be less than or equal to the base metal thickness (weld depth ⩽ base material thickness). Any sample violating this geometric rule was corrected by setting the weld depth equal to the base material thickness, a condition which represent the case of full penetration (burn-through).

**Preservation of Inter-Parameter Relationships:** To maintain the intrinsic correlations between process parameters, we implemented a series of co-adjustment rules. These rules ensured that perturbations to interdependent variables remained physically consistent, thereby preserving the underlying process dynamics. Key enforced relationships included:

Welding current ↑ → Welding duration ↓ : An inverse relationship was maintained.

Preheat temperature ↑ → Interpass temperature ↑ : A positive correlation was enforced.

Wire diameter ↑ → Welding current ↑ : To maintain arc stability, an increase in wire diameter triggered a commensurate increase in welding current.

Base material thickness ↑ → Wire diameter ↑ : Reflecting equipment compatibility and heat sink requirements.

Welding duration ↓ → Interpass temperature ↓ : Accounting for thermal dynamics, a shorter single pass duration led to a proportional decrease in the interpass temp.

**Parameter Range Clipping:** All parameters were clipped to their physically admissible operating ranges, which were determined based on the properties of the K439B superalloy and the actual possible manufacturing processes. The specific value ranges applied were as follows:

Welding current: 50-100 A

Weld depth: 0.5-20 mm

Weld area: 10-1000 mm²

Base material thickness: 1.5-20 mm

Single pass duration: 5-90 s

Interpass temperature: 20-800 °C

Cooling time (from welding completion to 500 °C): 3-15 s

Preheat temperature: 20-600 °C

Wire diameter: 0.8-3.2 mm

This multi-stage protocol, combining calibrated stochastic perturbation with deterministic physical reasoning, transformed the original 32 samples into a robust training ecosystem of 32,000 physically plausible data instances. This augmented dataset effectively captures the process variance within known physical bounds, providing a solid foundation for the subsequent discovery of explainable constitutive models.

**Engineering of Dimensionless Physical Terms and Symbolic Regression**

The discovery of an interpretable constitutive model was framed as a structured optimization problem, integrating symbolic regression with physics-guided feature engineering. The entire process was designed to efficiently navigate the vast space of possible equations and parameters. Our approach foundationally engineered six dimensionless physical terms ($T_1$-$T_6$) from domain expertise and first principles (Tables 2, 3), transforming raw parameters into physically meaningful quantities and constraining subsequent searches to plausible relationships. Each term was raised to a dimensionless exponent ($\alpha$, $\beta$, $\gamma$, etc.) to focus on functional relationships to the cracking probability index (CPI) without dimensional constraints.

A nested optimization architecture was implemented with two tiers. The outer loop employed Genetic Programming (GP) with a population size of 100 over 5 generations to explore equation structures using arithmetic operators and functions. For every candidate equation structure proposed by the GP, a full parameter optimization was performed using a Differential Evolution (DE) algorithm for 300 iterations (population size 50) with deliberately expanded parameter bounds covering the entire physically reasonable range (K: $1e^{-10}$-$1e^{-1}$; $\alpha,\beta,\gamma,\lambda,\mu,\nu$: 0.1-3.0; $\theta_1,\theta_2$: $1e^{-10}$-$1e^{-1}$; $\delta$: 1000-2000; k: 1.0-5.0). The algorithm automatically handled boundary constraints through built-in mechanisms. The fitness of each candidate model was quantified by a comprehensive accuracy metric: Fitness = 0.5 * Accuracy Validation Set + 0.5 * Accuracy Original 32 Samples. This ensured models performed well on both the augmented data and the ground-truth experiments.

Following the discovery of the optimal equation structure CPI = K * (($T_1$ * $T_4$) + ($T_2$ * $T_6$) + ($T_3$ * $T_5$)), we executed a rigorous refinement to maximize its predictive accuracy. The second local refinement phase utilized the L-BFGS-B algorithm for 5000 iterations, creating dynamic neighborhoods around the first-phase results. New bounds were defined as 0.1-10.0 times the initial optimized values while respecting the original global boundaries. Crucially, an adaptive boundary expansion mechanism was implemented: when the neighborhood range fell below 1% of the original parameter range, the algorithm automatically reverted to the original global bounds to ensure adequate search space.

**Hybrid Model Training and The Multi-Source Data Ecosystem**

To leverage the discovered constitutive equation for enhancing a high-performance predictive model, we constructed a hybrid modeling framework based on a multi-source data ecosystem. We amalgamated three distinct data streams, assigning differential weights to prioritize data fidelity and physical consistency. A pivotal application of the finalized constitutive equation was its use as a high-fidelity data generator.

Virtual Data Generation via Targeted Interpolation: The constitutive equation was deployed to perform targeted interpolation within the parameter space. It was specifically used to enrich regions that remained data-sparse even after the initial physics-informed augmentation. By systematically sampling input parameters and computing the corresponding CPI and crack probability, the equation generated an additional 32,000 virtual samples. The cracking probability for each virtual sample was calculated using the logistic function $P_{crack} = \frac{1}{(1+e^{(-k(CPI-CPI_{crit})})}$ with k=2 and $CPI_{crit}$=0.6, and samples with $P_{crack} > 0.5$ were labeled as "cracked" while others as "not cracked", ensuring physically consistent annotation throughout the dataset.

Deep Neural Network (DNN) Architecture and Training: The combined and weighted dataset was used to train a DNN with the following specifications to ensure robust learning:

Architecture: The model consisted of an input layer (9 neurons), two hidden layers (64 and 32 neurons, both with ReLU activation), and an output layer (1 neuron with sigmoid activation).

Regularization: To prevent overfitting, we incorporated L2 weight regularization and Dropout layers with rates of 0.3 and 0.2 after the first and second hidden layers, respectively.

Training Configuration: The model was compiled using the Adam optimizer with an initial learning rate of 0.001, which decayed exponentially. The loss function was binary cross-entropy. The model was trained for a maximum of 200 epochs with a batch size of 512.

Training Strategy: We employed a weighted loss function, where the sample weight for each data point was determined by its source (5.0 for original, 1.0 for augmented and virtual). Additionally, class weights were applied (1.5 for the minority "crack" class) to address label imbalance. Training incorporated early stopping (patience=20) to halt training if validation performance did not improve.

**Valdation and Reproducibility**

Model performance was evaluated using a held-out test set that was excluded from all development activities. Additional validation included physical consistency checks and sensitivity analysis to ensure robust performance across the parameter space. The complete analytical workflow was implemented in Python 3.8 using standard scientific computing libraries (NumPy, SciPy, scikit-learn, TensorFlow 2.8). Computational experiments were conducted only on a Lenovo ThinkBook 16+ laptop (equipped with an Intel Core Ultra5 processor, 32GB RAM, 1TB SSD, and NVIDIA RTX4060 graphics), with total computation time of approximately 1 hour for the complete framework execution. Code and processed data supporting the findings are available from the corresponding author upon reasonable request.

## SUPPLEMENTARY MATERIALS

This work was supported by Advanced Materials-National Science and Technology Major Project (No. 2025ZD0608901).